\documentclass[sn-mathphys,Numbered]{sn-jnl}
\usepackage{graphicx}%
\usepackage{multirow}%
\usepackage{amsmath,amssymb,amsfonts}%
\usepackage{amsthm}%
\usepackage{mathrsfs}%
\usepackage[title]{appendix}%
\usepackage{xcolor}%
\usepackage{textcomp}%
\usepackage{manyfoot}%
\usepackage{booktabs}%
\usepackage{algorithm}%
\usepackage{algorithmicx}%
\usepackage{algpseudocode}%
\usepackage{listings}%
\theoremstyle{thmstyleone}

\theoremstyle{thmstyletwo}%

\theoremstyle{thmstylethree}%

\begin{document}

\title[Thinking Twice: Clinical-Inspired Thyroid Ultrasound Lesion Detection Based on Feature Feedback]{Thinking Twice: Clinical-Inspired Thyroid Ultrasound Lesion Detection Based on Feature Feedback}

\author[1]{\fnm{Lingtao} \sur{Wang}}\email{hit\_wanglingtao@163.com}

\author*[1]{\fnm{Jianrui} \sur{Ding}}\email{jrding@hit.edu.cn}

\author[1]{\fnm{Fenghe} \sur{Tang}}\email{543759045@qq.com}

\author[2]{\fnm{Chunping} \sur{Ning}}\email{152081340@qq.com}

\affil[1]{\orgdiv{School of Computer Science and Technology}, \orgname{Harbin Institute of Technology}, \orgaddress{ \city{Harbin}, \country{China}}}

\affil[2]{\orgdiv{Ultrasound Department}, \orgname{The Affiliated Hospital of Qingdao University}, \orgaddress{\city{Qingdao}, \country{China}}}

\abstract{Accurate detection of thyroid lesions is a critical aspect of computer-aided diagnosis. However, most existing detection methods perform only one feature extraction process and then fuse multi-scale features, which can be affected by noise and blurred features in ultrasound images. In this study, we propose a novel detection network based on a feature feedback mechanism inspired by clinical diagnosis. The mechanism involves first roughly observing the overall picture and then focusing on the details of interest. It comprises two parts: a feedback feature selection module and a feature feedback pyramid. The feedback feature selection module efficiently selects the features extracted in the first phase in both space and channel dimensions to generate high semantic prior knowledge, which is similar to coarse observation. The feature feedback pyramid then uses this high semantic prior knowledge to enhance feature extraction in the second phase and adaptively fuses the two features, similar to fine observation. Additionally, since radiologists often focus on the shape and size of lesions for diagnosis, we propose an adaptive detection head strategy to aggregate multi-scale features. Our proposed method achieves an AP of 70.3\% and AP50 of 99.0\% on the thyroid ultrasound dataset and meets the real-time requirement. The code is available at \textcolor{magenta}{https://github.com/HIT-wanglingtao/Thinking-Twice.}}




\maketitle

\section{Introduction}\label{sec1}
\indent 
Thyroid nodules are a common condition with a high incidence rate \cite{bray2018globocan}, which can lead to thyroid dysfunction and even cancer. Palpation is often insufficient for detecting thyroid nodules, but with the increasing popularity of thyroid ultrasound, the detection rate has improved \cite{rago2008role}.

Ultrasound technology is widely used in clinical medical diagnostic tasks due to its low detection cost, real-time imaging, and non-invasive nature. It obtains images by receiving and processing reflected signals, allowing doctors to observe the range and physical properties of lesions in real-time. However, ultrasound technology has some limitations, including poor image quality, unclear lesion features, low resolution and high levels of noise. Therefore, the accuracy of ultrasound diagnosis is influenced by the clinician's experience and subjective factors \cite{khachnaoui2018review}. Computer-aided diagnosis can provide objective references for clinical diagnosis, improve the efficiency of clinicians' work, and reduce the number of missed diagnoses and misdiagnoses. In the clinical diagnosis process, physicians typically first identify the general lesion feature through rough observation and then focus on the lesion area to make a diagnosis based on detailed features and adjacent features of the lesion. Inspired by this diagnostic process, we propose a feature feedback mechanism for the one-stage lesion detection algorithm. In this mechanism, the feedback feature map with high semantic prior knowledge is obtained through feedback selection in the first feature extraction phase and used to enhance the attention of lesion features in the second feature extraction phase. To improve the detection head's ability to learn adjacent features and detailed shape features of multi-scale lesions, we propose an adaptive detection head based on a divide-and-conquer strategy, which performs divide-and-conquer preprocessing on multi-level features. By adding a weight-unshared preprocessing block to each layer, a single detection head can perform different preprocessing on multi-level features to improve the ability of adaptive spatial aggregation and long-distance dependency extraction for lesions of different sizes. The main contributions of this paper are as follows:
\begin{itemize}
\item
We applied the routine diagnostic process of physicians to the convolutional neural network and designed a feature feedback mechanism for one-stage ultrasound lesion detection.
\item
We proposed a feature pyramid based on the feature feedback mechanism and explored its effectiveness in low visual semantics and high visual semantics.
\item
We proposed an adaptive detection head based on a divide-and-conquer strategy to enhance the detection head's adaptability to learn shape features and adjacency features of multi-scale lesions.
\end{itemize}

\section{Related work}\label{sec2}
Yap et al. \cite{yap2017automated} compared deep learning-based breast ultrasound lesion detection methods to traditional detection methods relied on handcrafted features, and demonstrated the advantages of deep learning in this field. Current deep learning algorithms for lesion detection can be broadly classified into three groups: two-stage methods that combine candidate region with classification (e.g., R-CNN \cite{girshick2014rich} series), single-stage methods that transform target detection into a regression problem (such as the You Only Look Once (YOLO) \cite{redmon2016you} and Single Shot Detection (SSD) \cite{liu2016ssd} approaches), and Transformer-based target detection methods, such as end-to-end object detection with Transformers (DETR) \cite{carion2020end}. Li et al. \cite{li2018improved} used an improved Fast R-CNN model to detect papillary thyroid carcinoma and achieved a 93.5\% accuracy. Yap et al. \cite{yap2020breast} employed Faster R-CNN for breast ultrasound lesion detection and localization, achieving an F1 score of 93.2\%. Cao et al. \cite{cao2017breast} compared four deep learning models for breast cancer detection (Fast R-CNN, Faster R-CNN, YOLO and SSD) and concluded that SSD had the highest accuracy and recall. Chiang et al. \cite{chiang2018tumor} proposed a computer-aided detection system based on 3-D convolutional neural networks (CNNs) and prioritized candidate aggregation, achieved sensitivities of 95\%.

The two-stage algorithm suffers from high computational redundancy and slow detection speed, which cannot meet the real-time requirements of ultrasonic inspection. R-CNN \cite{girshick2014rich} was the first to use CNN for target detection tasks. SPPNet \cite{he2015spatial} proposed a spatial pyramid pooling layer to fuse multi-scale features, while Faster R-CNN \cite{ren2015faster} introduced a regional proposal network (RPN) to optimize the extraction of candidate boxes. DetectoRS \cite{qiao2021detectors} proposed Recursive Feature Pyramid (RFP) to enrich the expression ability of FPN \cite{lin2017feature} through a bottom-up backbone. However, the high computational redundancy of the two-stage algorithm makes it difficult to apply in real-time lesion detection.

Transformer-based target detection methods rely more on labeled data than CNN methods \cite{wang2022towards}, and they can show superior performance on data-rich datasets. However, with limited labeled data, Transformer-based detectors may have poor detection performance. DETR \cite{carion2020end} was the first to use Transformer for target detection. DAB-DETR \cite{liu2022dab} uses dynamic anchor coordinates as queries in Transformer decoder. DINO \cite{zhang2022dino} uses a contrastive way for denoising training and a mixed query selection method to initialize anchors, but its performance on small datasets is still limited.

The one-stage algorithm can meet the real-time requirements but is susceptible to ultrasound image noise. SSD \cite{liu2016ssd} achieves one-stage detection by combining prediction boxes from multiple non-fused feature maps. RetinaNet \cite{lin2017focal} addresses the detection performance issue caused by data imbalance using Focol loss. Yolov3 \cite{redmon2018yolov3} introduces multi-scale prediction and Logistic classifier, offering fast detection speed and strong versatility. Centernet \cite{duan2019centernet} and FCOS \cite{tian2019fcos} use centrality to suppress prediction boxes that deviate from the target's center, improving detection efficiency by eliminating anchors. Varifocalnet \cite{zhang2021varifocalnet} improves the detection head of FCOS to enhance the effect of dense object detection. Yolof \cite{chen2021you} only uses one layer of features of the backbone to achieve efficient target detection, but its performance on large targets is poor. Efficientdet \cite{tan2020efficientdet}, Yolox \cite{ge2021yolox} and Yolov7 \cite{wang2022yolov7} employ two-way feature fusion and feature reuse to enhance multi-level prediction.

Despite the use of advanced feature fusion techniques to enhance feature extraction capabilities, one-stage methods remain vulnerable to noise in ultrasound image detection due to their reliance on the traditional direct localization and classification mechanism. Inspired by clinical diagnostic workflows, our feature feedback mechanism performs a feedback operation on feedback-free features of the first feature extraction phase to achieve “think twice” process, as illustrated in Fig.1. Feedback-free features generate feedback feature maps through Selection module. These feedback feature maps are input into the backbone to guide the second feature extraction phase to generate feedback-based features.

Compared to the traditional mechanism, the feature feedback mechanism adds high-semantic prior knowledge to feature extraction, directing the CNN to focus more on the lesion area features rather than background noise. Compared to the traditional detection head, the adaptive head we proposed uses a weight-unshared preprocessing block to divide and conquer multi-level features to enhance adaptability to learn shape features and adjacency features of multi-scale lesions.

\begin{figure}[t]%
\centering
\includegraphics[width=\textwidth]{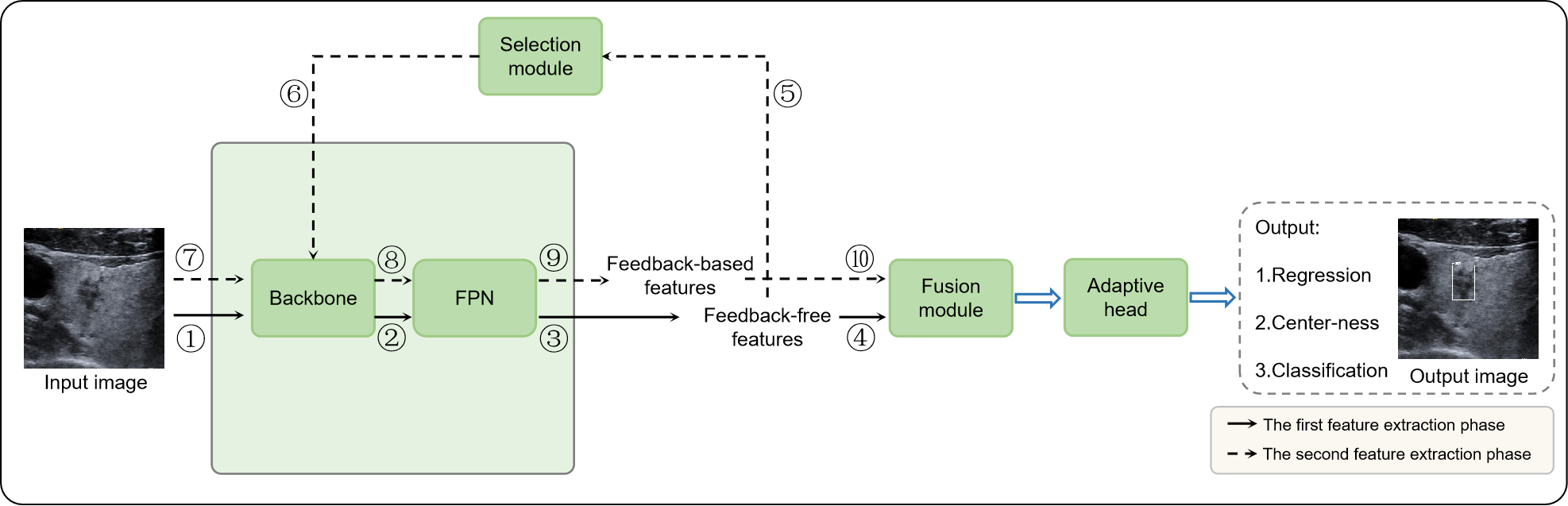}
\caption{Overview of our proposed feature feedback mechanism}\label{fig1}
\end{figure}

\section{Method}\label{sec3}
Our proposed method comprises three parts: the first feature extraction phase, the second feature extraction phase and adaptive detection, as illustrated in Fig.2. The input image passes through the backbone and Feature Pyramid Network (FPN) to obtain the initial feedback-free features $P_3^1$-$P_7^1$. Next, the feature selection module generates the feedback feature map $R_3$-$R_5$, which guides the second stage of feature extraction in the backbone to obtain the feedback-based features $P_3^2$-$P_7^2$. Finally, the two sets of output features ($P_3^1$-$P_7^1$ and $P_3^2$-$P_7^2$) are fused, and the adaptive detection head performs multi-level prediction to yield the lesion category, prediction boxes, and center-ness.

\begin{figure}[t]%
\centering
\includegraphics[width=\textwidth]{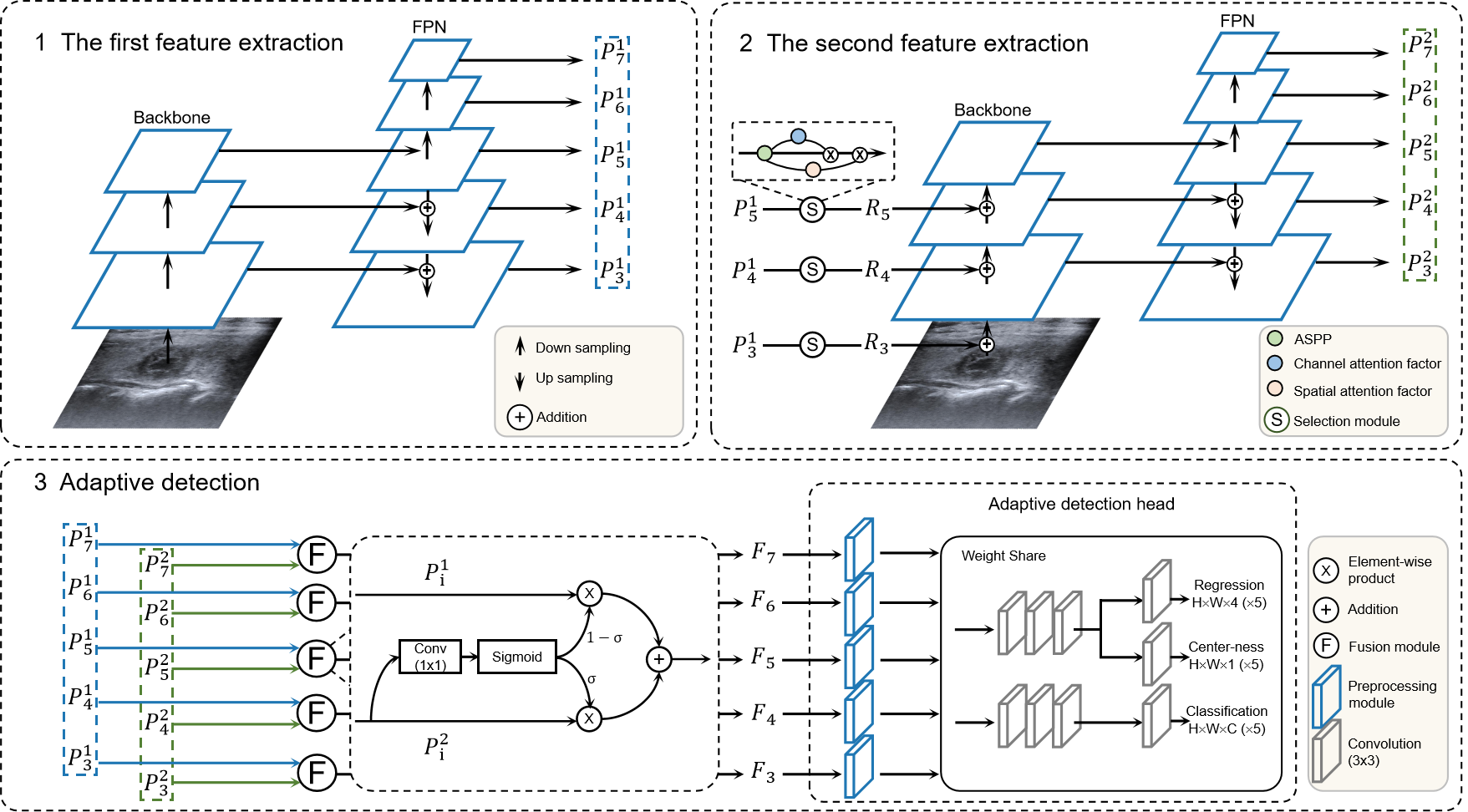}
\caption{Flowchart of our network}\label{fig2}
\end{figure}

\subsection{Feature Feedback Pyramid}\label{subsec1}
Ultrasound imaging often suffers from low resolution, resulting in blurred lesion features. To address this challenge, we introduce a feature feedback mechanism to the shallow layer of FPN to enhance the ability to extract lesion features in low-resolution images. The feature feedback mechanism filters features extracted in the first phase through a feature selection module, selectively enhancing and suppressing them with two learnable feature attention factors: $\sigma_1$ (channel attention factor) and $\sigma_2$ (spatial attention factor) to generate a feature map where the lesion area is enhanced. 

Benefiting from the prior knowledge of the feedback feature map, the lesion feature of the feedback-based feature is significantly enhanced, improving the ability of FPN to extract lesion features in low-resolution images. To prevent gradient disappearance and network degradation during model training, we use skip connections to fuse the initial feedback-free features and the feedback-based features for prediction.

The feature feedback mechanism expands the predicted feature from the original $P_i^1$ to $(1-w) \times P_i^1+w \times P_i^2$, where $w$ represents the selection weight generated by 1 $\times$ 1 convolution. If $w$ is 0, $P_i^1$ is used as the predictive feature, and if $w$ is not 0, the weighted sum of $P_i^1$ and $P_i^2$ is used instead. By fusing feedback-free features and feedback-based features, the fusion features used for prediction have stronger feature expressiveness.

Compared to traditional FPN, our proposed feature feedback pyramid incorporates the feature feedback mechanism at the low-level semantic layer ($P_3$, $P_4$, $P_5$) to improve local feature extraction capabilities. At the high-level semantic layer ($P_6$, $P_7$), Tang et al. \cite{tang2018pyramidbox} have shown that the feature fusion effect of FPN is poor. Therefore, we add $P_6$ and $P_7$, generated by down sampling $P_5$, to enrich the diversity of features.

As shown in Fig.2, the feature feedback pyramid incorporates feedback feature selection modules at the low-level semantic layer ($P_3$, $P_4$, $P_5$), performs feature feedback selection on feature $P_i^1$, extracted in the first phase, and then input it to the backbone to generate feature $P_i^2$ in the second phase, as shown in formula (1) - (3).
\begin{equation}
P_5^1=Conv\left(C_5^1\right),P_5^2=Conv\left(B_5\left(S\left(P_5^1\right),C_4^2\right)\right)\label{eq1}
\end{equation}
\begin{align}
P_4^1&=Conv\left(C_4^1\right)+Resize\left(P_5^1,2\right) \nonumber \\
P_4^2&=Conv\left(B_4\left(S\left(P_4^1\right),C_3^2\right)\right)+Resize\left(P_5^2,2\right)\label{eq2}
\end{align}

\begin{align}
P_3^1&=Conv\left(C_3^1\right)+Resize\left(P_4^1,2\right) \nonumber\\
P_3^2&=Conv\left(B_3\left(S\left(P_3^1\right),C_2^2\right)\right)+Resize\left(P_4^2,2\right)\label{eq3}
\end{align}
where $P_i^j$ denotes the $j$-th phase output feature of the FPN $P_i$ layer, $C_i^j$ is the $j$-th phase output feature of the backbone $C_i$ layer, $Conv$ represents the convolutional operation with a kernel of 1, $Resize(.,r)$ denotes upsampling or downsampling with a sampling rate $r$, $S$ is the feedback feature selection operation (explained in Section 3.2), and $B_i$ represents the calculation of the $i$-th stage of the backbone (described in Section 3.3).

At the high semantic level, $P_6$ and $P_7$ are obtained by downsampling $P_5$ as formula (4) and (5).
\begin{equation}
P_6^1=Resize\left(P_5^1,1/2\right),P_6^2=Resize\left(P_5^2,1/2\right) 
\end{equation}
\begin{equation}
P_7^1=Resize\left(P_6^1,1/2\right),P_7^2=Resize\left(P_6^2,1/2\right) 
\end{equation}

Finally, use the fusion module (F module in Fig.2) to fuse the two output features for prediction as formula (6).
\begin{equation}
 F_i=P_i^1\times\left(1-\sigma\left(Conv\left(P_i^2\right)\right)\right)+P_i^2\times\sigma\left(Conv\left(P_i^2\right)\right) 
\end{equation}
where $F_i$ is the fusion feature, $\sigma$ is the Sigmoid function.

\subsection{Feedback feature selection module}\label{subsec2}
To suppress noise and extract valuable lesion features from ultrasound images, the feedback feature selection module employs several techniques, namely Atrous Spatial Pyramid Pooling (ASPP) \cite{chen2017deeplab}, channel attention factor $\sigma_1$, and spatial attention factor $\sigma_2$, for multi-scale feature fusion and selection. ASPP combines both global and local information at multiple scales through image pooling and dilated convolution to capture context and semantic information effectively. This makes it easier to extract multi-scale lesion information while ignoring noise and local texture. The channel attention factor suppresses noise through global hybrid pooling and generates selection weights for each channel using the pooled value. The spatial attention factor captures long-range spatial dependencies through depthwise convolution and generates spatial selection weights that rely on high semantic features. The structure of feedback feature selection module is illustrated in Fig.3.
\begin{figure}[t]%
\centering
\includegraphics[width=0.9\textwidth]{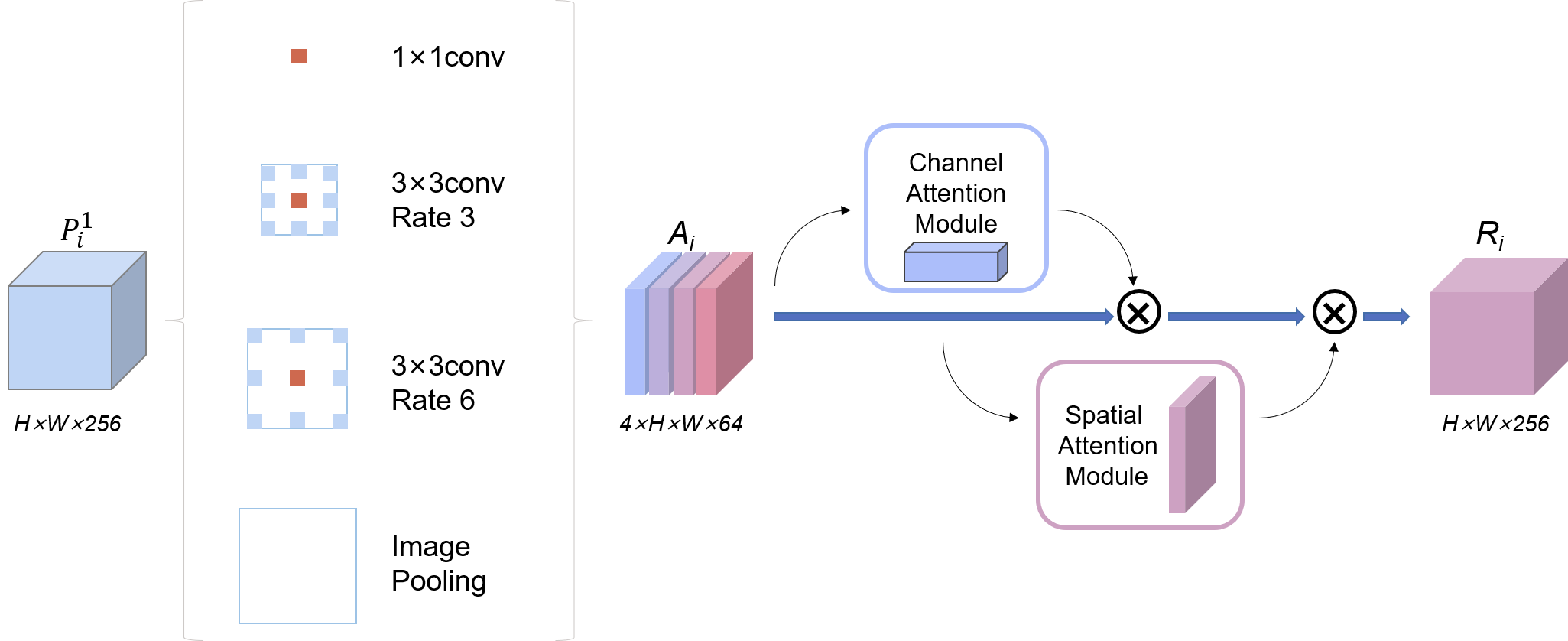}
\caption{Feedback feature selection module}\label{fig3}
\end{figure}

As shown in Fig.3, the module is implemented by first using the four branches of ASPP (the convolution branch with a kernel size of 1, the dilated convolution branch with dilated rate of 3 and 6 respectively, and the image pooling branch) to generate four feature maps with a channel number of C/4. These four feature maps are concatenated to obtain multi-scale features $A_i$. Then two parallel branches are employed to calculate channel feature attention and spatial feature attention, resulting in the generation of $\sigma_1$ and $\sigma_2$. The channel attention module comprises a hybrid pooling layer and a convolution operation with a kernel size of 1. The spatial attention module comprises a depth wise convolution operation. $A_i$ is multiplied by $\sigma_1$ and $\sigma_2$ to yield $R_i$.

If the input feature is $P_i^1$, ASPP, $\sigma_1$, $\sigma_2$ and $R_i$ are calculated as formula (7) - (9).

\begin{equation}
A_i=Concat\left(Conv\left(P_i^1\right),Conv\left(P_i^1,r=3\right),Conv\left(P_i^1,r=6\right),AvgPool\left(P_i^1\right)\right)
\end{equation}
\begin{equation}
\sigma_1=\sigma\left(Conv\left(AvgPool\left(A_i\right)+MaxPool\left(A_i\right)\right)\right), \sigma_2=\sigma\left(Depthwise C o n v\left(A_i\right)\right)
\end{equation}
\begin{equation}
R_i=A_i\times\sigma_1\times\sigma_2  
\end{equation}
where $A_i$ represents the output feature of ASPP, $Conv\left(.,r\right)$ is the dilated convolution with a dilation rate of $r$, $AvgPool$ is the average pooling operation, $MaxPool$ is the maximum pooling operation, $DepthwiseConv$ is the depthwise convolution with a kernel size of 7, $R_i$ is the output feature of the module.
\subsection{Improvement of the backbone}\label{subsec3}
While most algorithms adopt ResNet \cite{he2016deep} and ResNeXt \cite{xie2017aggregated} as the backbone network, we employ the more advanced ConvNext \cite{liu2022convnet} in our approach. ConvNext expands the receptive field and network width through the design of depthwise convolution and inverted bottleneck blocks, which enhances the backbone network's ability to extract global features of lesions. To accommodate the feature feedback mechanism, we have made improvements to the backbone as illustrated in Fig.4.

When the feedback features $R_i$ are input to the backbone, $R_i$ branches are added to the $C_3$-$C_5$ layers of the backbone to facilitate feedback operations. Point convolution is employed to ensure that the number of channels of $R_i$ is equivalent to the number of channels in down-sampled feature maps $C_{i-1}^2$. The accumulated features are then fed into the Convnext blocks to generate $C_i^2$, as expressed by formula (10).
\begin{equation}
C_i^2=B_i\left(Conv\left(R_i\right)+Resize\left(LN\left(C_{i-1}^2\right),1/2\right)\right)
\end{equation}
where $B_i$ represents the calculation of N ConvNext blocks ($C_3$-$C_5$ layers have N values of 3, 9 and 3 respectively), $Resize(.,1/2)$ denotes down sampling, and $LN$ represents layer normalization.
\begin{figure}[t]%
\centering
\includegraphics[width=0.4\textwidth]{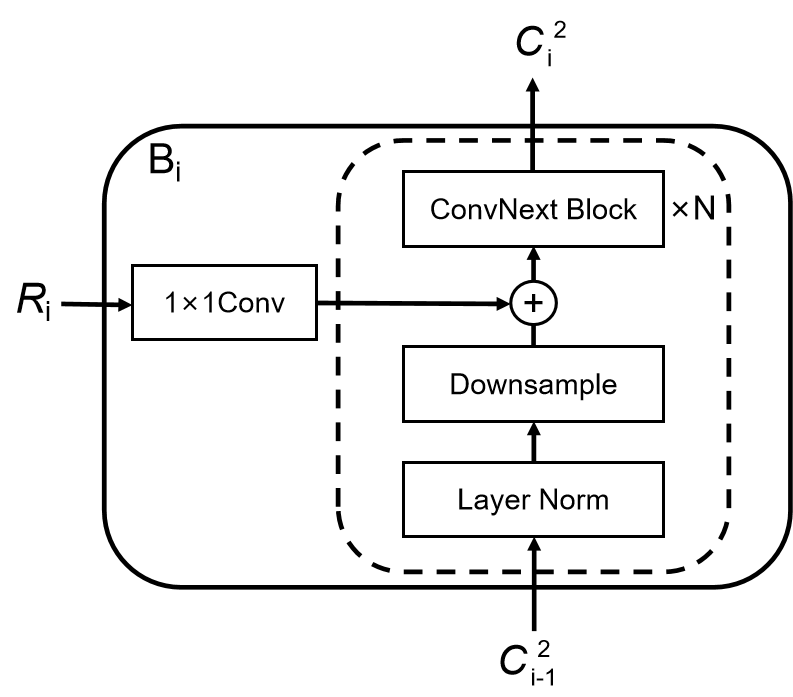}
\caption{Improved backbone with ConvNext}\label{fig4}
\end{figure}
\subsection{Adaptive detection head}\label{subsec4}
To detect lesion objects of varying sizes in ultrasound images, we propose an adaptive detection head to enhance its ability to adapt to multi-scale lesion features, as illustrated in Fig. 5. The weight-sharing detection head uses the same weights for multi-level features. However, multi-level features correspond to lesions of varying sizes and shapes, posing a challenge during the weights learning process. Therefore, we incorporate a weight-unshared preprocessing block before the weight-sharing detection head to enhance its ability to handle multi-level features.

The convolution used by the traditional detection head struggles to capture the long-distance dependencies in the image and cannot perform adaptive spatial aggregation on the lesion area. Deformable convolution \cite{dai2017deformable,zhu2019deformable} has better adaptive spatial aggregation ability. However, the free offset of convolution points (the blue points of the deformable convolution in Fig. 5) may result in an offset between the center of the receptive field before deformation and the center of the receptive field after deformation. An anchor point may also not be included in the deformed receptive field when predicting its bounding box.

Ultrasound lesion is mostly characterized by aggregated lump-like nodule, with fuzzy and irregular spreading areas around the main nodule. In order to better extract lesion shape features, we propose a deformable surround convolution that redesigns the deformation mode and scale of deformable convolution. The adaptive feature preprocessing block we propose combines deformable surround convolution and depth wise separable convolution to enhance adaptive spatial aggregation ability for lump-like lesions and focus on fuzzy adjacency features.

As illustrated in Fig.5, the deformable surround convolution fixes the center convolution point and expands the surround points outward. Each surround point learns an offset that does not exceed the maximum threshold and offsets according to fixed direction to achieve adaptive learning of lesion shape. The preprocessing block reduces the difficulty of learning lump-like features and ensures focus on the center of lesion.

The detection head’s calculation process changes from $H(F_i)$ to $H(w_i(F_i))$, where $w_i$ is learnable preprocessing block and $H$ is a single detection head with weight sharing. Compared to directly detecting $F_i$ of different feature levels, $w_i$ enhances the detection head's fitting ability to $F_i$.
\begin{figure}[t]%
\centering
\includegraphics[width=1\textwidth]{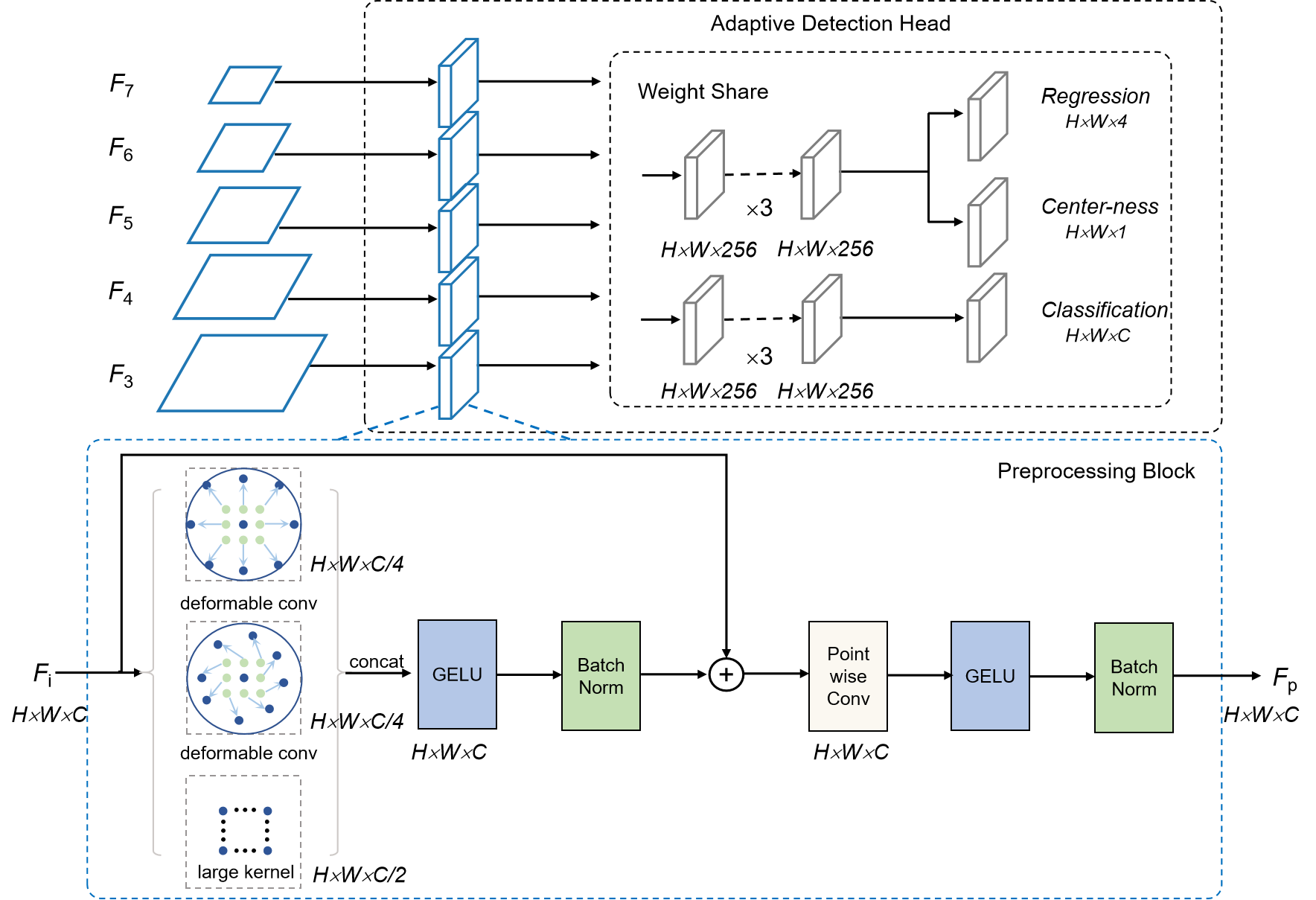}
\caption{Structure of adaptive detection head}\label{fig5}
\end{figure}

\begin{figure}[b]%
\centering
\begin{minipage}[b]{0.3\textwidth}
  \centering
  \centerline{\includegraphics[width=1\textwidth]{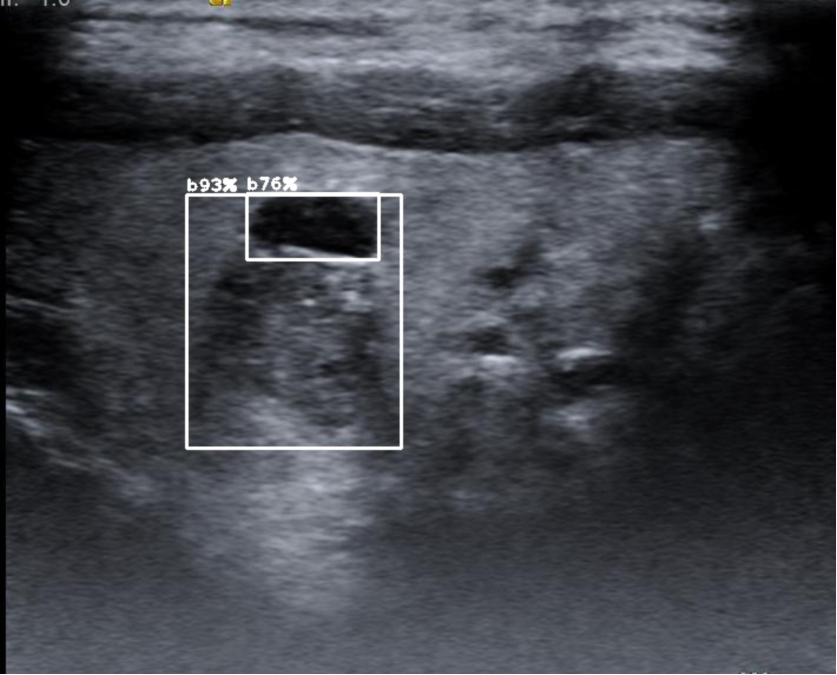}}
  \centerline{\scriptsize{(a)Without preprocessing block}}
\end{minipage}
\hspace{.35in}
\begin{minipage}[b]{0.3\textwidth}
  \centering
  \centerline{\includegraphics[width=1\textwidth]{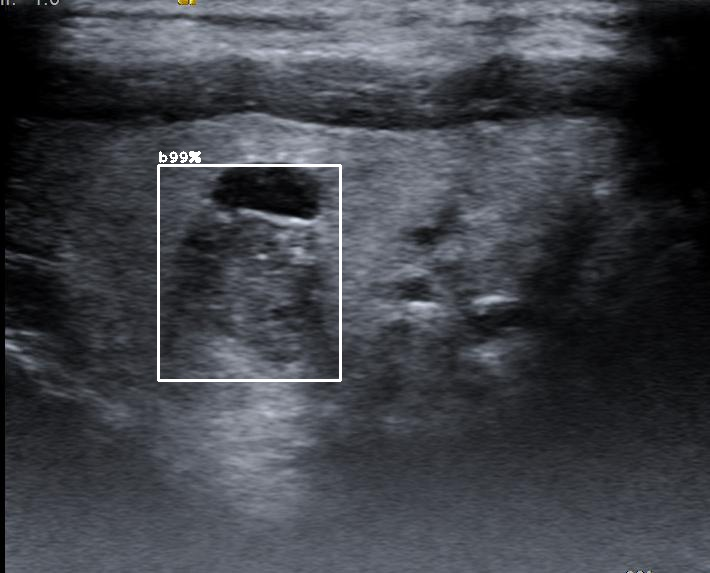}}
  \centerline{\scriptsize{(b)With preprocessing block}}
\end{minipage}
\caption{Example image of overlapping lesions}\label{fig6}
\end{figure}

In certain cases, regions of large lesions may comprise small lesions with strong features and sprawling areas with weak features. Since the weight-sharing detection head employs the same convolution weights for different semantic layers, it can prioritize detecting lesion areas of corresponding size at different semantic layers while disregarding other features such as sprawl with larger areas but weak features. This can lead to the detection of small lesions in isolation or the overlapping of large and small lesions. Fig.6 illustrates this phenomenon, which is mitigated by the addition of preprocessing blocks. 
\section{Experimentation}\label{sec4}
\subsection{Datasets}\label{subsec1}
The dataset used in this study, provided by the Affiliated Hospital of Qingdao University, consists of 1023 annotated thyroid ultrasound images, each with a size of approximately 573$\times$710 pixels. The images were acquired using an HIVSION 900 ultrasound scanner and include Region of Interest (ROI) annotations by physicians, along with corresponding diagnosis results. Fig.7 illustrates some cases of the dataset.

\begin{figure}[h]%
\centering
\begin{minipage}[h]{0.24\textwidth}
  \centering
  \centerline{\includegraphics[width=1\textwidth]{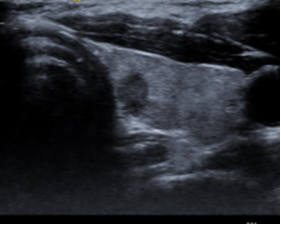}}
  \centerline{\scriptsize{(a)Malignant case\ \ }}
\end{minipage}
\begin{minipage}[h]{0.24\textwidth}
  \centering
  \centerline{\includegraphics[width=1\textwidth]{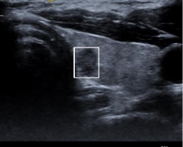}}
  \centerline{\scriptsize{(b)Malignant case with ROI}}
\end{minipage}
\begin{minipage}[h]{0.24\textwidth}
  \centering
  \centerline{\includegraphics[width=1\textwidth]{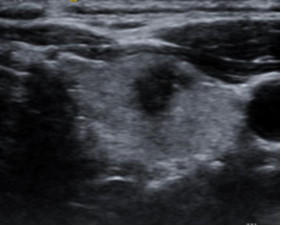}}
  \centerline{\scriptsize{(c)Benign case}}
\end{minipage}
\begin{minipage}[h]{0.24\textwidth}
  \centering
  \centerline{\includegraphics[width=1\textwidth]{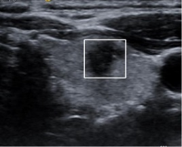}}
  \centerline{\scriptsize{(d)Benign case with ROI\ \ }}
\end{minipage}
\caption{Examples of the dataset}\label{fig7}
\end{figure}
\subsection{Experiment Details}\label{subsec2}
We adopt FCOS as our baseline model and employ the SGD optimizer in our experiments. The initial learning rate is set to 0.01, with a momentum of 0.9, weight decay of 0.0001, batch size of 4, and a total of 50,000 training steps. We apply a learning rate decrease by a factor of 0.1 at 25,000 and 35,000 steps. The dataset is randomly split into 60\% for training, 20\% for validation, and 20\% for testing. Input images are resized to 800×1024 and augmented with random flipping. The loss function used for training is defined by formula (11).
\begin{align}
L\left(\{c_{x,y}\},\{t_{x,y}\},\{o_{x,y}\}\right)=\frac{1}{N_{pos}}\{&\sum_{x,y}{L_{cls}\left(c_{x,y},c_{x,y}^\ast\right)}+\sum_{x,y}I_{\{c_{x,y}^\ast>0\}}L_{reg}\left(t_{x,y},t_{x,y}^\ast\right)  \nonumber \\
+&\sum_{x,y}I_{\{c_{x,y}^\ast>0\}}L_{ctn}\left(o_{x,y},o_{x,y}^\ast\right)\}
\end{align}
where $c_{x,y}$, $t_{x,y}$, $o_{x,y}$ represent the predicted category, box, and center-ness, respectively, for point (x, y), and $c_{x,y}^\ast$, $t_{x,y}^\ast$, $o_{x,y}^\ast$ represent their corresponding ground truth values. $N_{pos}$ is the number of positive samples. $L_{cls}$ is the focal loss. $L_{reg}$ is the IOU loss , and $L_{ctn}$ is the center-ness loss, which is calculated using the binary cross-entropy function (BCE). $I_{\{c_{x,y}^\ast>0\}}$ is an indicator function that evaluates to 1 if the predicted category for (x, y) is positive, and 0 otherwise.

During training, input images are resized to 800$\times$1024 and fed to the ConvNeXt backbone network in batches to extract features $C_3^1$, $C_4^1$, $C_5^1$. FPN then performs the first feature fusion to obtain $P_3^1$, $P_4^1$, $P_5^1$, as well as down sampling $P_5^1$ to generate $P_6^1$, $P_7^1$, resulting in the first set of output feature maps. The feedback feature selection module uses $P_3^1$, $P_4^1$, and $P_5^1$ to generate feedback features $R_3$, $R_4$, and $R_5$, which guide the second feature extraction process in the backbone network, resulting in the second set of feature maps $P_3^2$-$P_7^2$. The fusion module then combines the corresponding features from the two sets $P_i^1$ and $P_i^2$ to obtain the fusion feature $F_i$. Finally, the adaptive detection head performs multi-scale prediction on the fusion feature, outputting predicted category, boxes, and center-ness. We calculate the prediction loss and update the model parameters accordingly.

During evaluation, the test dataset images are fed one-by-one into the trained model to obtain predicted category, box, and center-ness for each image. Non-maximum value suppression is applied to remove redundant prediction boxes generated during the detection process, producing a set of high-quality prediction boxes, which are then visualized. To evaluate the detection performance, we use the pycocotools target detection evaluation tool to compare predicted results with manually annotated ground truth. The evaluation metrics include average precision (AP), AP at 50\% IoU overlap (AP50), and AP at 75\% IoU overlap (AP75), which are calculated using the following equations (12)-(14):
\begin{equation}
AP=\frac{1}{MR}\sum_{m=1}^{M}\sum_{r=0}^{R-1}\sum_{k=1}^{N}{max_{\hat{k}\geq k}P_{Iou>0.5+0.05r}\left(\hat{k}\right)\Delta r_{Iou>0.5+0.05r}(k)}
\end{equation}
\begin{equation}
AP50=\frac{1}{M}\sum_{m=1}^{M}\sum_{k=1}^{N}{max_{\hat{k}\geq k}P_{Iou>0.5}\left(\hat{k}\right)\Delta r_{Iou>0.5}\left(k\right)}
\end{equation}
\begin{equation}
AP75=\frac{1}{M}\sum_{m=1}^{M}\sum_{k=1}^{N}{max_{\hat{k}\geq k}P_{Iou>0.75}\left(\hat{k}\right)\Delta r_{Iou>0.75}\left(k\right)}
\end{equation}
where $M$ is the number of categories, $R$ is the number of IoU thresholds, and $N$ is the number of predicted instances. $P_{IoU>a}$ and $r_{IoU>a}$ represent the precision and recall rates respectively, when the IoU threshold is $a$. $\Delta r$ denotes the change in recall rate as the threshold varies, and $max_{\hat{k}\geq k}P\left(\hat{k}\right)$ represents the maximum precision at each recall threshold.
\begin{table}[t]
\caption{Comparison of detection accuracy of thyroid ultrasound lesions (\%)}\label{tab1}
\renewcommand\arraystretch{1.3}
\setlength\tabcolsep{8pt}
\begin{tabular}{llccccc}
\toprule
Method              & Backbone & AP   & AP50 & AP75 & AP\tiny{benign} & AP\tiny{malignant} \\ \midrule
Faster RCNN \cite{ren2015faster} & Resnet50 & 64.3 &96.6      & 79.2     & 61.5         & 67.1            \\
               RetinaNet \cite{lin2017focal}    &   Resnet50   &  65.2    &   97.6   & 80.3         & 62.4    &  67.9      \\
             Yolov3 \cite{redmon2018yolov3}       &    Darknet53      &   64.7   &95.2      &81.5      &62.5          &66.8             \\
                FCOS \cite{tian2019fcos}    & Resnet50         &65.8      &95.5      & 80.8     & 63.5         & 68.2            \\
 EfficientDet \cite{tan2020efficientdet}                  &EfficientNet-B1          &66.1      & 98.7     & 77.1     &63.8          &68.5             \\
    VarifocalNet \cite{zhang2021varifocalnet}                & Resnet50         & 64.5     &  97.3    & 78.5     &  64.4        & 64.6            \\
  Yolof \cite{chen2021you}                  &   Resnet50       & 65.9     & \bf{99.2}     & 81.4     & 64.8         &  66.9           \\
 Yolox \cite{ge2021yolox}                  & Darknet53         &67.0      & 98.1     & 83.4     & 64.4         &69.5             \\ 
Yolov7 \cite{wang2022yolov7}                  & CBS+ELAN    &67.3    & 98.3    &84.0    & 65.3       &69.2       \\\midrule
   DETR \cite{carion2020end}                 & Resnet50         &63.4      & 93.6     &76.2      &61.2         & 65.7            \\
 DAB-DETR \cite{liu2022dab}                 &Resnet50          &64.9      &96.3      &78.9      &64.1          &65.8             \\
 DINO \cite{zhang2022dino}                   &Resnet50          &66.1      &95.8      &83.6      & 62.5         &69.7             \\ \midrule
 Ours                   &Resnet50          &69.6      &99.0      &    87.7  &68.2          & 71.0            \\
 Ours                   &  Convnext-tiny        &\bf{70.3}      & 99.0     &\bf{88.4}      &\bf{68.9}          & \bf{71.6}            \\ \botrule
\end{tabular}
\end{table}
\begin{figure}[t]%
\centering
\begin{minipage}[h]{0.19\textwidth}
  \centering
  \centerline{\includegraphics[width=1\textwidth]{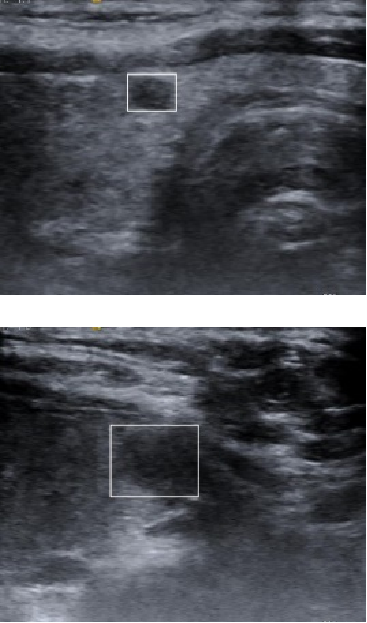}}
  \centerline{\scriptsize{(a)Ground Truth}}
\end{minipage}
\begin{minipage}[h]{0.19\textwidth}
  \centering
  \centerline{\includegraphics[width=1\textwidth]{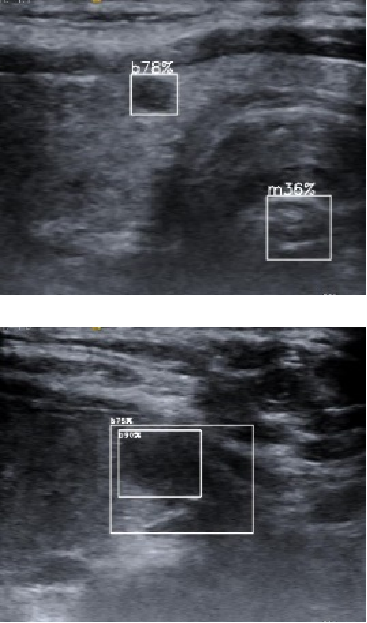}}
  \centerline{\scriptsize{(b)FCOS}}
\end{minipage}
\begin{minipage}[h]{0.19\textwidth}
  \centering
  \centerline{\includegraphics[width=1\textwidth]{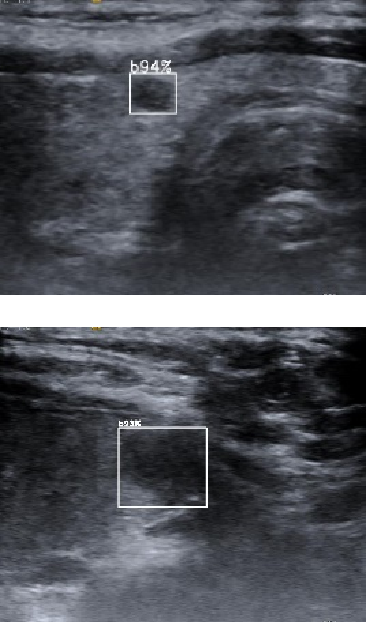}}
  \centerline{\scriptsize{(c)Yolov7}}
\end{minipage}
\begin{minipage}[h]{0.19\textwidth}
  \centering
  \centerline{\includegraphics[width=1\textwidth]{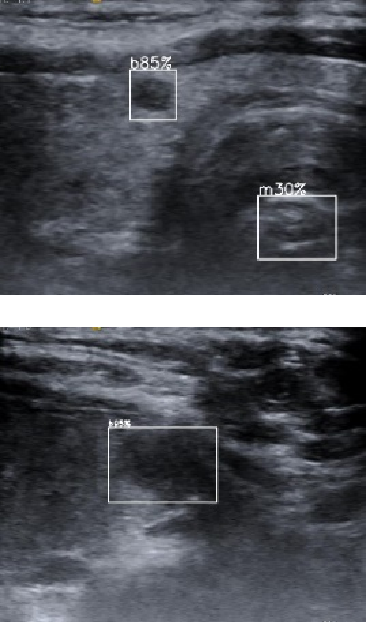}}
  \centerline{\scriptsize{(d)DINO}}
\end{minipage}
\begin{minipage}[h]{0.19\textwidth}
  \centering
  \centerline{\includegraphics[width=1\textwidth]{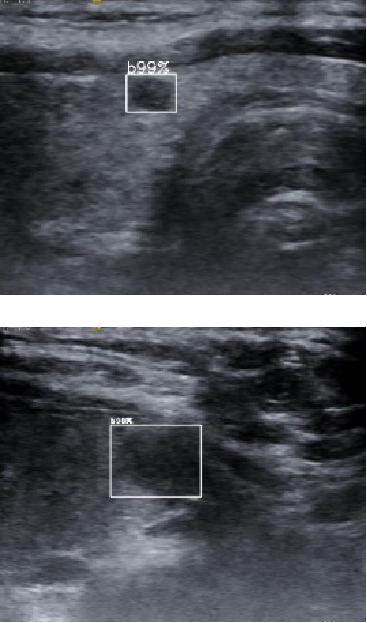}}
  \centerline{\scriptsize{(e)Ours}}
\end{minipage}
\caption{Examples of lesion detection results}\label{fig7}
\end{figure}
\subsection{Comparative experiment}\label{subsec3}
We conducted a comparative experiment between our algorithm and mainstream algorithms on the thyroid ultrasound dataset, as shown in Table 1. When using backbones of the same size, compared with Faster RCNN \cite{ren2015faster}, RetinaNet \cite{lin2017focal}, Yolov3 \cite{redmon2018yolov3}, FCOS \cite{tian2019fcos}, and VarifocalNet \cite{zhang2021varifocalnet} that use one-way fusion FPN, with significantly improved AP by 6.0\%, 5.1\%, 5.6\%, 4.5\%, and 5.8\%, respectively. This is because their one-way fusion FPN is susceptible to noise and cannot fully extract lesion features, and their backbone networks have poor global feature extraction ability. Transformer-based DETR \cite{carion2020end}, DAB-DETR \cite{liu2022dab}, and DINO \cite{zhang2022dino} are not only slow in convergence speed but also difficult to leverage Transformer performance advantages on small datasets. Our method achieves 4.2\% higher AP than DINO, demonstrating superior performance than Transformer-based detectors on small-scale ultrasound datasets. Yolof \cite{chen2021you} only uses a single-level feature for prediction, resulting in poor multi-level prediction performance. EfficientDet \cite{tan2020efficientdet} uses BiFPN to achieve efficient two-way feature fusion, while Yolox \cite{ge2021yolox} and Yolov7 \cite{wang2022yolov7} use PAFPN to enhance the ability of feature reuse and two-way feature fusion. The two-way fusion method improves by about 1.5\% compared to the one-way fusion method, but is still 3\% lower than our method using the feature feedback mechanism. This is because recursive computation with high semantic feedback features has better feature extraction capability than single two-way feature fusion computation. In addition, the ConvNeXt architecture we use also has better global feature extraction ability to improve detection accuracy.

Fig.8 illustrates the lesion detection results of the compared algorithms. DINO and FCOS exhibit lesion overlap and false detection. In contrast, Yolov7 and our algorithm achieve better detection results, with our algorithm showing greater precision.

\subsection{Ablation experiment}\label{subsec4}
We conducted ablation experiments on the thyroid ultrasound dataset using FCOS as the baseline model, as shown in Table 2. The adding of ConvNeXt resulted in a 1.7\% increase in detection AP compared to ResNet, demonstrating the improved feature extraction ability of ConvNeXt. The adaptive detection head further improved detection AP by 1\%, indicating that the weight-unshared preprocessing block enhances the fitting ability of different levels of features. Finally, the addition of the feature feedback pyramid led to a significant 1.8\% improvement in detection AP, demonstrating the enhancement of the extraction ability of local lesion features through the feedback mechanism.

We compared the baseline detection head (FCOS) with the coupling detection head (Yolov3), the decoupling detection head (Yolox), and the adaptive detection head, as shown in Table 3. The coupling detection head only uses one branch to perform regression and classification tasks, leading to conflicts between different tasks and resulting in a 1.5\% lower detection AP value than decoupled structures. Both the baseline detection head and the decoupling detection head use a decoupled structure, suppressing prediction boxes that deviate from the target by predicting the center-ness and IoU scores, respectively. As a result, their detection performance is similar. Our adaptive detection head adds a weight-unshared preprocessing module to each layer of features, enhancing detection performance on multi-level features and achieving the highest AP value for lesion detection.

We conducted detection precision comparison and real-time verification experiments on FPN without feedback, FPN with $P_3$-$P_5$ feedback, and FPN with $P_3$-$P_7$ feedback, as shown in Table 4. The results indicate that FPN with feedback achieves significantly higher detection precision than FPN without feedback, and the feedback feature selection module effectively improves detection precision. Adding feedback in the low semantic layer produces a more noticeable effect than in the high semantic layer, which we attribute to the high semantic layer already having a large receptive field. When we add a feedback feature map with a higher receptive field, the receptive field is already much larger than the size of the lesion, rendering the addition of a feedback feature map to the high semantic layer unnecessary. In terms of detection speed, the simple calculation process of one-stage algorithms allows the feedback methods to meet the real-time requirements of ultrasonic detection (Scanning imaging speed higher than 24 frames per second can realize real-time imaging \cite{eidheim2005real}).

\begin{table}[t]
\caption{Ablation test of lesion detection precision (\%)}\label{tab2}
\renewcommand\arraystretch{1.3}
\setlength\tabcolsep{20pt}
\begin{tabular}{p{4cm}ccc}
\hline
Method              & AP   & AP50 & AP75 \\ \hline
Baseline & 65.8 & 95.5     &  80.8    \\
+convnext          & 67.5    &98.7      &84.8      \\
+convnext+adhead    &  68.5    &  98.6    & 86.8     \\
+convnext+adhead+FB-FPN   &70.3      &99.0      &88.4      \\ \hline
\end{tabular}
\end{table}
\begin{table}[t]
\caption{Comparison of different detection heads (\%)}\label{tab3}
\renewcommand\arraystretch{1.3}
\setlength\tabcolsep{20pt}
\begin{tabular}{p{4cm}ccc}
\hline
Method              & AP   & AP50 & AP75 \\ \hline
Baseline head (FCOS) & 67.5 & 98.7    &  84.8    \\
Coupling head (Yolov3)         & 65.6    &97.1      &82.0      \\
Decoupling head (Yolox)    &  67.1    &  98.5    & 87.0     \\
Adaptive head (Ours)   &68.5      &98.6     &86.8     \\ \hline
\end{tabular}
\end{table}
\begin{table}[t]
\caption{Comparison of different feedback methods}\label{tab4}
\renewcommand\arraystretch{1.3}
\setlength\tabcolsep{8pt}
\begin{tabular}{p{5cm} cccc}
\hline
Method              & AP(\%)   & AP50(\%) & AP75(\%) & FPS\\ \hline
feedback-free & 68.5 & 98.6    & 86.8  & 46  \\
$P_3$-$P_5$feedback+ASPP        & 69.6    &98.6      &87.4   & 40   \\
$P_3$-$P_7$feedback+ASPP    &  69.6    &  98.4   & 87.8  & 34  \\
$P_3$-$P_5$feedback+ASPP+$\sigma_1$+$\sigma_2$ (Ours)   &70.3      &99.0     &88.4 & 39  \\
$P_3$-$P_7$feedback+ASPP+$\sigma_1$+$\sigma_2$   &  70.1    &  98.5   & 88.2 & 30  \\ \hline
\end{tabular}
\end{table}
\begin{figure}[H]
\centering
\begin{minipage}[h]{0.22\textwidth}
  \centering
  \centerline{\includegraphics[width=1\textwidth]{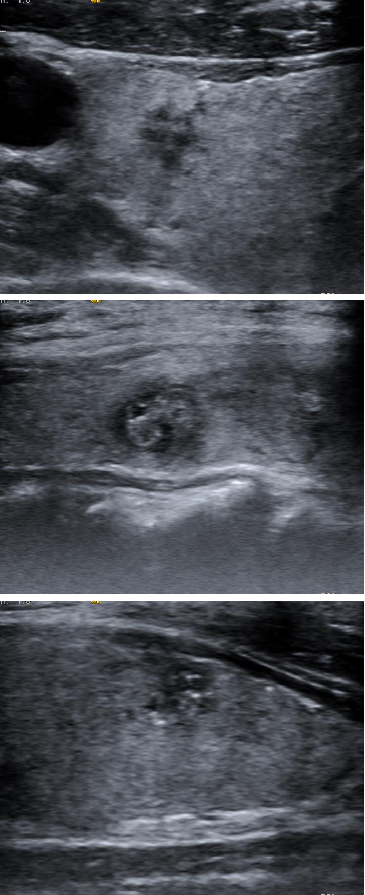}}
  \centerline{\scriptsize{(a)Original image}}
\end{minipage}
\begin{minipage}[h]{0.22\textwidth}
  \centering
  \centerline{\includegraphics[width=1\textwidth]{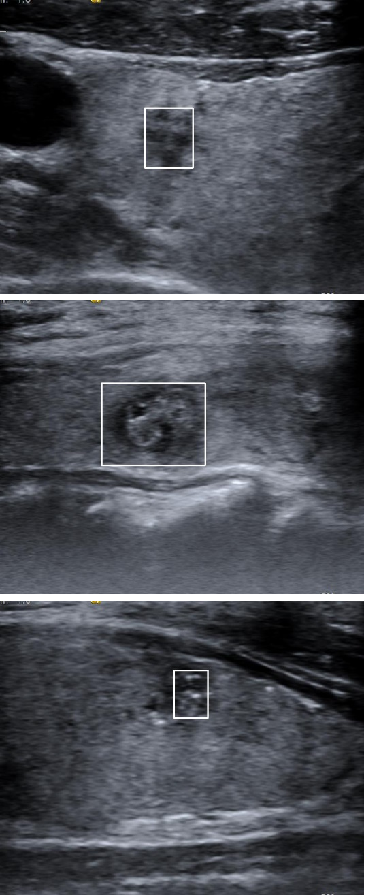}}
  \centerline{\scriptsize{(b)Ground Truth}}
\end{minipage}
\begin{minipage}[h]{0.22\textwidth}
  \centering
  \centerline{\includegraphics[width=1\textwidth]{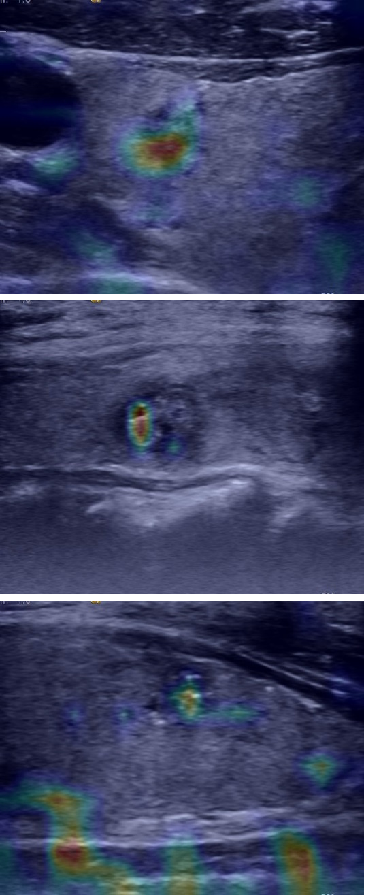}}
  \centerline{\scriptsize{(c)Without feedback}}
\end{minipage}
\begin{minipage}[h]{0.22\textwidth}
  \centering
  \centerline{\includegraphics[width=1\textwidth]{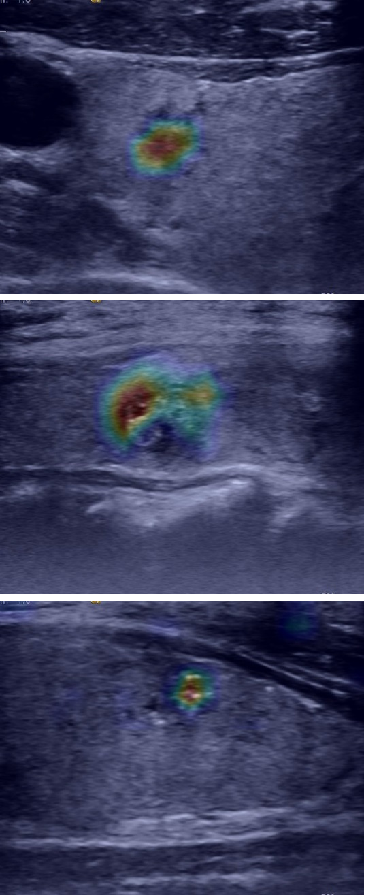}}
  \centerline{\scriptsize{(d)With feedback}}
\end{minipage}
\caption{Visualize with Gradient thermodynamic}\label{fig9}
\end{figure}
\subsection{Visualization experiment}\label{subsec5}
We used Grad-CAM \cite{selvaraju2017grad} to visualize the attention areas of lesions before and after adding the feature feedback mechanism, as shown in Fig.9. After adding feature feedback, the points of interest in the background are suppressed, and the degree of attention in the lesion area is enhanced.

To visualize the changes in data distribution during model training, we conducted visualization experiments using the pre-trained mapping layer to simulate feature maps of the model, as shown in Fig.10. Compared to the first phase output feature maps ($P_3^1$,$P_4^1$,$P_5^1$), the second phase output feature maps ($P_3^2$,$P_4^2$,$P_5^2$) exhibit reduced redundancy and effectively suppressed noise. This demonstrates that feature feedback selection has the effect of suppressing local noise.

\begin{figure}[h]%
\centering
\includegraphics[width=1\textwidth]{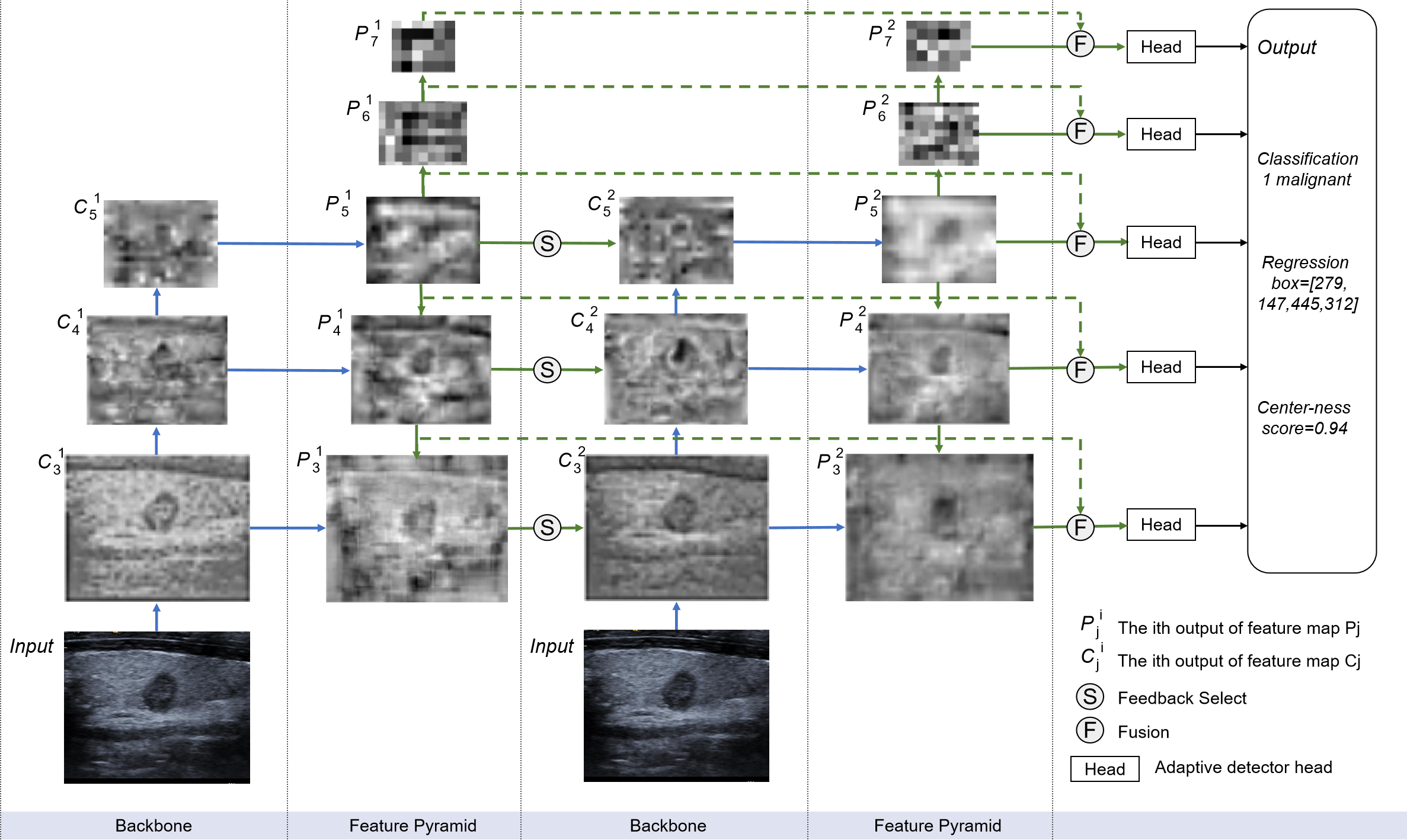}
\caption{Characteristic map simulation example}\label{fig10}
\vspace{-0.4cm}
\end{figure}
\subsection{Other experiment}\label{subsec6}
To verify the generality of our method, we conducted a comparative experiment on the breast ultrasound dataset BUSI \cite{al2020dataset}. BUSI contains 647 annotated images with widths ranging from 324 to 719 and heights ranging from 190 to 1048, as shown in Fig.11.

The BUSI dataset suffers from a severe sample imbalance problem, with 437 benign samples and only 210 malignant samples. After randomly dividing the dataset into training, validation, and test sets, only around 120 malignant samples are available for training. Moreover, many malignant samples exhibit blurred spread areas and exceed the boundary of the image, as illustrated in Fig.11. These factors lead to low overall detection accuracy in breast malignancy samples, as shown in Table 5. Nonetheless, our method achieves higher detection accuracy than other methods, demonstrating its versatility in the field of ultrasound.
\begin{figure}[h]%
\centering
\begin{minipage}[h]{0.24\textwidth}
  \centering
  \centerline{\includegraphics[width=1\textwidth]{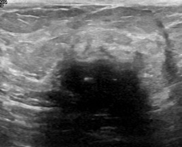}}
  \centerline{\scriptsize{(a)Malignant case}}
\end{minipage}
\begin{minipage}[h]{0.24\textwidth}
  \centering
  \centerline{\includegraphics[width=1\textwidth]{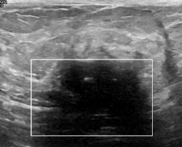}}
  \centerline{\scriptsize{(b)Malignant labeled case}}
\end{minipage}
\begin{minipage}[h]{0.24\textwidth}
  \centering
  \centerline{\includegraphics[width=1\textwidth]{fig11c.png}}
  \centerline{\scriptsize{(c)Benign case}}
\end{minipage}
\begin{minipage}[h]{0.24\textwidth}
  \centering
  \centerline{\includegraphics[width=1\textwidth]{fig11d.png}}
  \centerline{\scriptsize{(d)Benign labeled case}}
\end{minipage}
\caption{Examples of BUSI}\label{fig11}
\vspace{-0.7cm}
\end{figure}

\begin{table}[h]
\caption{Comparison of detection accuracy of breast ultrasound lesions (\%)}\label{tab5}
\renewcommand\arraystretch{1.3}
\setlength\tabcolsep{8pt}
\begin{tabular}{llccccc}
\toprule
Method              & Backbone & AP   & AP50 & AP75 & AP\tiny{benign} & AP\tiny{malignant} \\ \midrule
Faster RCNN \cite{ren2015faster} & Resnet50 & 42.1 &66.4      & 45.0     & 53.6         & 30.6            \\
               FCOS \cite{tian2019fcos}  &   Resnet50   &  43.4    &   67.3   & 46.2         & 56.0    &  30.8      \\
            Yolov7 \cite{wang2022yolov7}       &   CBS+ELAN    &  46.5   &68.1   &51.8     &\bf{59.9}        &33.1            \\
     DETR \cite{carion2020end}    & Resnet50         &41.6      &66.1      & 40.0   &53.6      & 29.7      \\
DINO \cite{zhang2022dino}                  &Resnet50     &44.8      & \bf{72.4}    &46.5    &52.8        &36.8         \\
Ours              & Convnext-tiny       &\bf{49.1}   & 68.9   & \bf{60.1}    &  59.4       & \bf{38.9}           \\ \botrule
\end{tabular}
\end{table}

\section{Conclusion}\label{sec5}
In conclusion, this work proposes a one-stage ultrasound lesion detection algorithm with a feature feedback mechanism and a detection head adaptive strategy. Inspired by the clinical diagnosis process of making a rough observation followed by a detailed observation of lesion features, our algorithm implements a ``thinking twice'' process that extracts high semantic prior knowledge and uses it to guide the second feature extraction. The detection head adaptive strategy enhances the algorithm's ability to identify lesions of different sizes and spreading areas. Our algorithm achieves superior performance on the thyroid ultrasound dataset while meeting real-time requirements.

However, this work has some limitations. The proposed algorithm still faces challenges in detecting small and low-contrast lesions due to the limitations of ultrasound imaging technology. Additionally, the proposed algorithm's performance on the BUSI dataset, which suffers from sample imbalance and other challenges, shows that there is still room for improvement in the generalizability of the algorithm.

Future work could involve improving the algorithm's ability to detect small and low-contrast lesions and enhancing its generalizability to other ultrasound datasets with varying challenges. Additionally, exploring the potential of the proposed ``thinking twice'' process and adaptive feature preprocessing block in other medical imaging fields and natural image detection could be an exciting direction for future research.

\bmhead{Acknowledgments}
This work is supported by Shandong Natural Science Foundation of China (ZR2020MH290) and by the Joint Funds of the National Natural Science Foundation of China (U22A2033).

\section*{Declarations}
\bmhead{Conflict of Interest}
The authors declared that they have no conflicts of interest to this work.

\bibliography{sn-bibliography}% common bib file
\end{document}